\documentclass[sn-mathphys,Numbered]{sn-jnl}% Math and Physical Sciences Reference Style
\usepackage{CJKutf8}
\usepackage{graphicx}%
\usepackage{multirow}%
\usepackage{amsmath,amssymb,amsfonts}%
\usepackage{amsthm}%
\usepackage{mathrsfs}%
\usepackage[title]{appendix}%
\usepackage{xcolor}%
\usepackage{textcomp}%
\usepackage{manyfoot}%
\usepackage{booktabs}%
\usepackage{algorithm}%
\usepackage{algorithmicx}%
\usepackage{algpseudocode}%
\usepackage{listings}%
\usepackage{natbib}
\usepackage{xspace}
\newcommand{\method}{DoubleCheck\xspace}
\theoremstyle{thmstyleone}%
\theoremstyle{thmstyletwo}%

\theoremstyle{thmstylethree}%

\begin{document}

\title[LTCR: \textbf{L}ong-\textbf{T}ext \textbf{C}hinese \textbf{R}umor Detection Dataset]{LTCR: \textbf{L}ong-\textbf{T}ext \textbf{C}hinese \textbf{R}umor Detection Dataset}


\author[1]{\fnm{Ziyang} \sur{Ma}}\email{mazy23@mail2.sysu.edu.cn}
\equalcont{These authors contributed equally to this work.}

\author[1]{\fnm{Mengsha} \sur{Liu}}\email{liumsh6@mail2.sysu.edu.cn}
\equalcont{These authors contributed equally to this work.}

\author[1]{\fnm{Guian} \sur{Fang}}\email{fanggan@mail2.sysu.edu.cn}

\author*[1]{\fnm{Ying} \sur{Shen}}\email{sheny76@mail.sysu.edu.cn}

\affil[1]{\orgdiv{School of Intelligent Systems Engineering}, \orgname{Sun Yat-Sen University}}


% \abstract{Healthcare rumors spread fast on social media, negatively influencing the citizens' mindset for healthcare and their countermeasures to diseases. To better detect all of the fake news, especially long texts which are harder to find completely, a \textbf{L}ong-\textbf{T}ext \textbf{C}hinese \textbf{R}umor detection dataset named LTCR is proposed. The dataset consists of 1,729 and 500 pieces of real and fake news, respectively. The average lengths of real and fake news are approximately 230 and 152 characters. We also propose \method, Salience-aware Fake News Detection Model, which achieves the highest accuracy (95.85\%), fake news recall (90.91\%) and F-score (90.60\%) on the dataset.\footnote{https://github.com/Enderfga/DoubleCheck}}

\abstract{False information can spread quickly on social media, negatively influencing the citizens' behaviors and responses to social events. To better detect all of the fake news, especially long texts which are harder to find completely, a \textbf{L}ong-\textbf{T}ext \textbf{C}hinese \textbf{R}umor detection dataset named LTCR is proposed.
The LTCR dataset provides a valuable resource for accurately detecting misinformation, especially in the context of complex fake news related to COVID-19.
The dataset consists of 1,729 and 500 pieces of real and fake news, respectively. The average lengths of real and fake news are approximately 230 and 152 characters. We also propose \method, Salience-aware Fake News Detection Model, which achieves the highest accuracy (95.85\%), fake news recall (90.91\%) and F-score (90.60\%) on the dataset\textsuperscript{1}\footnote[0]{\textsuperscript{1}https://github.com/Enderfga/DoubleCheck}.}




\maketitle

\section{Introduction}\label{sec1}
% 第一层（为什么重要）：论证中国社交媒体上新冠假新闻的传播会对 Healthcare 造成负面影响

%% 1. 阐述 Fake News on social media 对 Healthcare 的负面影响，以论证假新闻检测对于 Healthcare 的重要性
%%    - fake cures 关于治疗措施的假新闻会误导大众对新冠预防、诊断、治疗等的观念，例如``喝酒精可以治疗新冠``
%%    - false conspiracy theories 关于新冠起源的阴谋论会造成群众甚至医护人员的恐慌，对 mental health 有害
%% 2. 因此，假新闻的传播会造成广泛的社会影响，全球假新闻的数量大、传播后造成的影响广泛（列举数据），因此医疗假新闻的检测十分重要；
% Rumors impact people's behaviors and attitudes toward many events, such as the coronavirus disease 2019 (COVID-19)~\cite{intro-impacts}. Especially, fake news on social media spreads quickly and broadly on the Internet, misleads public opinion, and is harmful to properly fighting against severe diseases. 
% As illustrated in Table~\ref{data-example}, fake cures such as ``Drinking alcohol can cure COVID-19'' may encourage patients to attempt them, which is even worse for their health than the virus itself~\cite{intro-fake-cures}. 
% On the other hand, fake conspiracy theories, partly manipulated by social bots~\cite{intro-fake-conspiracy}, threaten the mental health of patients and medical workers. For example, the theory saying that the technology used in the vaccine could alter people’s DNA frightens lots of people and leads to protests against the vaccination~\cite{intro-dna-conspiracy}. 
% Therefore, the existence and spreading of fake medical information on social media greatly hurt healthcare. Detecting healthcare rumors on social platforms is quite important. To narrow down the range of our study, COVID-19 rumor detection is chosen as the main research field.

Rumors can have a significant impact on people's behaviors and attitudes\cite{intro-impacts}, especially in the era of social media, where fake news can spread rapidly and mislead public opinion. 
The COVID-19 pandemic has only exacerbated this problem, as illustrated in Table~\ref{data-example}, fake cures such as ``Drinking alcohol can cure COVID-19'' may encourage patients to attempt them, which is even worse for their health than the virus itself~\cite{intro-fake-cures}. 
On the other hand, fake conspiracy theories, partly manipulated by social bots~\cite{intro-fake-conspiracy}, leading to public confusion and potentially harmful outcomes. For example, the theory saying that the technology used in the vaccine could alter people’s DNA frightens lots of people and leads to protests against the vaccination~\cite{intro-dna-conspiracy}. 
Therefore, detecting and addressing rumors on social media platforms is an urgent issue. In this study, we focus on developing a dataset for detecting rumors in long Chinese texts, with a specific emphasis on verifying COVID-19 related misinformation.

\begin{table}[h] 
\caption{Examples of healthcare rumors relevant with fake cures and fake conspiracy theories.\label{data-example}}
\begin{CJK}{UTF8}{gbsn}

\begin{tabular}{c|l}
\hline 
\multirow{2}{*}{\text {Fake Cure}} 
& \text{\textbf{Text}: 喝酒能够治愈新冠。}\\
& \text{\textbf{Translation}: Drinking alcohol can cure COVID-19.}  \\
 \hline
\multirow{2}{*}{\text {Fake Conspiracy Theory }} & {\text{\textbf{Text}: 
疫苗中使用的技术会改变人的DNA。}} \\
& \text {\textbf{Translation}: The technology used in vaccine could alter people’s DNA.} \\
 \hline
\end{tabular}
\end{CJK}

\end{table}

% 第二层（其他人做了什么、存在什么问题）
%% 1. 科学问题:(1)假新闻短，长文本假新闻的查全不够好 (2)假新闻少，对假新闻检测能力的评估不准确
Benefiting from deep learning, more NLP (Natural Language Processing) researchers are interested in fake news detection, and the task is mainly regarded as a text classification task~\cite{ravichandran2023classification}. Many studies have been conducted on English long-text fake news detection, including datasets and models~\cite{cui2020coaid,cheng2021covid,hossain2020covidlies}. However, few works target that in the Chinese language. In the field of Chinese fake news detection, we argue the existence of problems below: (1) \textbf{For model training}, due to the fact-checking websites' condensing to original fake news, the lack of long-text fake news hampers the ability of trained models to find long-length fake news as ultimately as possible. (2) \textbf{For model evaluation}, a benchmark dataset called "the Chinese COVID-19 Fake News Dataset"(CHECKED)~\cite{yang2021checked} has been published in previous research. However,it has a limited number of fake news pieces (344) compared to real ones (1,760), leading to that the evaluation metrics mainly reflect the competence of detecting real news but mismatch that of detecting fake news.

% limited number of and 
% is not prioritized, leading to low fake news recalls and making it more possible for fake news to be leaked out in real applications.
We evaluate text classification models on short and long news of CHECKED~\cite{yang2021checked}, which reveals that the models have lower accuracy and recall on long-length news than short items. More details will be shown in Section~\ref{subsec-pretest}. \textit{A conclusion can be drawn that it is harder to find out all the fake news in long texts than in short texts. High-quality long-text datasets for Chinese fake news detection and better models are required to solve the problem.} 

\textbf{To alleviate the problem of missing high-quality datasets with enough long-length fake news for model training and evaluation}, we propose LTCR, \textbf{L}ong-\textbf{T}ext  \textbf{C}hinese \textbf{R}umor detection dataset. Specifically, we investigate available data sources as many as possible. Two kinds of sources are considered, which are formal and informal. Formal sources are datasets proposed in published papers. Informal sources include datasets published without authorized scripts and fact-checking websites where rumors are explained. Crawled data and existing datasets are combined as the dataset LTCR. 

% \begin{itemize}
%     % 缺乏一手的假新闻来源
%     \item \textbf{The lack of first-hand sources of Chinese fake news on COVID-19.} The main sources are those government-supported rumor refutal websites, where rumors have been condensed to be more reader-friendly and thus not original. Meanwhile, the rumors posted on these websites are not that ample to support data mining. Hence, the dataset is required to utilize as many sources as possible to guarantee its quality.
%     % 时效性
%     \item \textbf{The timeliness of existed COVID-19 Chinese fake news dataset.} Most existing formal or informal datasets merely contain data from the year 2020 and are not up-to-date. However, fake news can spread quickly and may become outdated within a short period of time. The dataset is required to have fake news in more recent years, making it more representative of the current state of fake news on COVID-19. 
% \end{itemize}

Moreover, \textbf{to validate the usability of the LTCR dataset and promote the ability of models to identify as much fake news as possible}, we propose Salience-aware Fake News Detection Model, \method. Salience means the importance of every token in a text. This model first detects the salient parts in the text, which contain more important information, and then extracts the feature from these vital parts to classify the news. Extensive experiments are carried out. We conduct comparative tests that demonstrate the effectiveness of \method\ by comparing it with baseline models, and ablation studies to validate the ability of the Input Re-weight module. The results show that our model has a better ability to detect fake news as completely as possible with good overall accuracy and precision, with the designed module providing much ability.

The main contributions of this paper are as follows:
\begin{enumerate}
    % 数据集
    \item[(1)] To alleviate the impact of the lack of long-length fake news in the Chinese language on model training and evaluation, we propose a Long-text Chinese Rumor dataset. Our dataset has been validated in the context of detecting COVID-19 rumors. Compared to existing benchmarks, our dataset includes a larger number of fake news samples, and importantly, features longer fake news samples to provide a more realistic and challenging evaluation scenario for models. 

    % 模型与数据集配套使用
    \item[(2)] Most of the current rumor detection models are designed for short news texts. There are very few models that specifically target long fake news texts, and their detection accuracy is not high. Therefore, we design a salience-aware fake news detection model to validate the usability of our dataset, which is suitable for long-text fake news detection.
    %and also inspire the further improvement of fake news detection models.

    % 我们比别人好
    \item[(3)] Our proposed dataset stands out from Chinese rumor dataset in terms of long-length news contents. We investigate the model performances by comparing with different baseline classifiers and conducting ablation analysis, revealing that our method outperforms others on the proposed dataset.
\end{enumerate}

The rest of the article is organized as follows. Section~\ref{sec2} reviews relevant work on Chinese fake news datasets and fake news detection methods. Section~\ref{sec3} describes our Long-text Chinese Rumor detection dataset (LTCR) and the Salience-Aware Fake News Detection model \method. Section~\ref{sec4} provides data statistics of our dataset LTCR. Section~\ref{sec6} shows experimental results and related analyses. Section~\ref{sec6} presents our discussion on our proposed LTCR and \method. Finally, in Section~\ref{sec7}, we make a conclusion of our work and point out several promising directions for future researches.

\section{Related Work}\label{sec2}
\subsection{Fake News Detection Datasets}
% 这段是抄10.1109/ICICS55353.2022.9811124这篇综述的（page3左栏），需要转述                                  
Studies present that there are many English datasets and few Chinese ones of long-text fake news detection~\cite{related-work-survey}. 
Table~\ref{datasets} shows some fake news datasets related to COVID-19 . CoAID is an English dataset released by Cui and Lee containing 926 social media posts from different sources, and they classified them as fake or true~\cite{cui2020coaid}. Hossain released COVIDLies dataset containing 6,761 tweets, and they annotated them into agree, disagree, or no Stance manually~\cite{hossain2020covidlies}. Cheng et al. ~\cite{cheng2021covid} collect rumors from various sources, including news sites, Fact-Checked websites, and Twitter; they get 6,834 posts and manually label them as true, false, or unverified. In addition, a dataset that includes 155,468 tweets is released by Micallef et al. and are classified into misinformation, counter misinformation, or irrelevant~\cite{micallef2020role}. 
Targeting Chinese datasets, CHECKED collects 344 fake posts and 1,760 real posts with their reposts and comments on Weibo from December 2019 to August 2020. It has a class-unbalanced problem because it has much less fake news than real news. To the best of our knowledge, no existing works focus on the problem of fully detecting long-text COVID-19 fake news. 

\begin{table}[h]
\caption{Fake News Detection Datasets.}\label{datasets}%
\begin{tabular}{@{}cccc@{}}
\toprule
\textbf{Dataset}	& \textbf{Size}	& \textbf{Language}   & \textbf{Source}	\\
\midrule
CoAID~\cite{cui2020coaid}		&   926 posts	& English  &  Facebook, Twitter, Instagram, etc.\\%Youtube, and TikTok
Cui and Lee~\cite{cheng2021covid}    &   6,834 posts  &   English & Twitter and Websites\\
COVIDLies~\cite{hossain2020covidlies}   &   6,761 tweets & English   & Twitter\\
Cheng et al.~\cite{micallef2020role}    & 155,468 tweets & English   & Twitter\\
CHECKED~\cite{yang2021checked} 	& 	2,104 posts		& Chinese  & Weibo	\\
\botrule
\end{tabular}

\end{table}

\subsection{Fake News Detection Methods}
%% 2. 社交媒体假新闻检测的主流方法：基于语义的方法、基于传播的方法、基于知识的方法
%%    - 分别介绍三种方法的原理、优点、缺点
%%    - 最后落脚到基于知识的方法可解释性更强、且适合假新闻的早期检测
Various approaches have been proposed for detecting fake news, including those based on semantics, propagation, and knowledge. They all regard rumor detection as a text classification task. 

\textbf{Propagation-based methods} analyze the propagation patterns of news on social media to identify patterns associated with fake news. A classic example is the (user)-(Twitter)-(news event) network, where Gupta et al.~\cite{2012Evaluating} determined the credibility relationship between these three entities to judge the authenticity of the news. Recently, Huang et al.~\cite{2020Deep} have provided a more fine-grained perspective, which treats the spatial structure and the temporal structure as a whole to model message propagation.
However, these methods are useful for social media posts that have a clear propagation history, but they may struggle with new posts that have no forwarding relationships.

\textbf{Knowledge-based methods} use external knowledge sources to verify the veracity of the news. 
To assess the authenticity of news articles, we need to compare the knowledge extracted from to-be-verified news content with the facts in a pre-built news database. Trivedi~\cite{2018LinkNBed} used entity resolution techniques to identify proper matchings.
This method has the advantage of being able to leverage external knowledge and provide evidence of the judgment, but it requires access to a large and up-to-date news database which costs a lot of storage resources.

\textbf{Semantics-based methods} extract the vocabulary, syntax, and semantic features of news content to identify patterns of fake news.
Machine Learning (ML) models have been mainly used for semantics-based fake news detection. For example, semantics-based methods have relied on SVMs~\cite{V2017Automatic}, Random Forests (RF)~\cite{2019Fake}, and XGBoost~\cite{2016XGBoost}.
Within the deep learning (DL) framework, there are many classic models used to extract potential text features of news, such as Text-CNN ~\cite{2014Convolutional}, LSTMs~\cite{1997Long}, GRUs~\cite{2014Learning}; and the Transformer~\cite{2018BERT}.
Recent researchers have made improvements based on traditional models.
Jain et al.~\cite{0CanarDeep} proposed a hybrid deep learning model based on attention, which utilizes Hierarchical Attention Network (HAN) and Multi-Layer Perceptron (MLP) to learn features for detecting rumor posts.
Shi et al.~\cite{2023PL}proposed a POS-aware and layer ensemble transformer neural network by incorporating the parts-of-speech type of words to learn more distinct word features for text classification.
Nevertheless, they are more focused on improving comprehensive accuracy, ignoring that recall is more important for fake news detection because missing fake news is riskier than missing real news. Pre-trained models such as BERT~\cite{2018BERT} and RoBERTa~\cite{liu2019roberta} are also used to get the context-dependent word or document embedding on the fake news detection task~\cite{kula2021implementation,jain2022aenet}. They usually have better performances, but require much more computational resources to do inferences. Different from previous work, our model is designed with the practical need to detect all fake news in mind, with low computational cost and enhanced performance for long texts.

\section{Methodology}\label{sec3}

\subsection{Schema Definition}
Fake news detection is to classify a news item as \textit{fake news or real news}. Providing high-quality long-text training and evaluating data is one of our core contributions, so other classes such as \textit{ambiguous news} is not focused. Some examples containing fake and real news are presented in Table~\ref{fake-real-example}. They are selected from our proposed dataset LTCR which will be mentioned in Section~\ref{LTCR}.

\begin{table}[h] 
\caption{Examples of fake news and real news.\label{fake-real-example}}
\begin{CJK}{UTF8}{gbsn}
\begin{tabular}{c|l}
\hline 
\multirow{8}{*}{\text {Fake News}} 
& \parbox[c]{11cm}{\textbf{Title}: 武汉三民小区的人全变红码，全市所有人全部核酸检测。}\\
& \parbox[c]{11cm}{\textbf{Translation}: All people in Sanmin Community in Wuhan have a red code, and all people in the city are tested for nucleic acid.}\\
& \parbox[c]{11cm}{\textbf{Summary}: 三民小区封掉，5000人都变红码动不了，又隔离14天，上班的也不行了。}\\
& \parbox[c]{11cm}{\textbf{Translation}: The Sanmin community was closed, and 5,000 people became red codes and could not move. They were quarantined for another 14 days, and those who went to work could not.}\\
&\parbox[c]{11cm}{\textbf{Text}: \textcolor{red}{武汉三民小区的人全变红码，全市所有人全部核酸检测！}武汉三民小区出了六个感染者，清零很久以后出的，震动全国。是一个89岁的，之前感染自己好了，没去治疗，隔了两个月又发病传染人。\textcolor{red}{武汉发了狠，所有人（除六岁以下）全部强制检测！}三民小区封掉，5000人都变红码动不了，又隔离14天，上班的也不行了。全市10天都测一遍，不测的人变黄码无法出行，过期检测自己出钱。要测一个小区，就提前封闭。这是搞毛了，被病毒害的太惨。大样本检测，科研意义也很大。}\\
& \parbox[c]{11cm}{\textbf{Translation}: \textcolor{red}{All people in Wuhan Sanmin District have a red code, and all people in the city have been tested for nucleic acid! }Wuhan’s Sanmin community had six infected people, and it was a long time after they were cleared, which shocked the whole country. It was an 89-year-old man who recovered from the infection before, but did not go for treatment, and then became ill and infected people again after two months. \textcolor{red}{Wuhan has become ruthless, and everyone (except those under the age of six) is forced to test! }The Sanmin community was closed, and 5,000 people became red codes and could not move. They were quarantined for another 14 days, and those who went to work were also unable to move. The whole city is tested once every 10 days, and those who fail to pass the yellow code are unable to travel, and they have to pay for the expired test. To measure a community, it is closed in advance. This is messed up, it was too badly harmed by the virus. Large-sample testing is also of great scientific significance.}  \\
&\parbox[c]{11cm}{\textbf{Time}: 2020/5/13 10:44:00}\\
 \hline
\multirow{8}{*}{\text {Real News}} 
& \parbox[c]{11cm}{\textbf{Title}: 巴基斯坦外长接机中国专家医疗队。巴铁，我们来了。}\\
& \parbox[c]{11cm}{\textbf{Translation}: The Pakistani foreign minister picked up the Chinese expert medical team. Our friends, here we come.}\\
& \parbox[c]{11cm}{\textbf{Summary}: 当地时间28日下午，中国援助巴基斯坦的8人专家医疗队抵达巴基斯坦首都伊斯兰堡。}\\
& \parbox[c]{11cm}{\textbf{Translation}: On the afternoon of the 28th local time, the 8-member expert medical team assisted by China arrived in Islamabad, the capital of Pakistan.}\\
&\parbox[c]{11cm}{\textbf{Text}: 当地时间28日下午，中国援助巴基斯坦的8人专家医疗队抵达巴基斯坦首都伊斯兰堡。巴基斯坦外长库雷希等多位政府官员在停机坪等候。随着专机带来的物资包括10万只医用口罩、医用防护服、核酸检测试剂等。}\\
& \parbox[c]{11cm}{\textbf{Translation}: The Pakistani Foreign Minister picked up the Chinese expert medical team Pakistan Railway, here we come! On the afternoon of the 28th local time, the 8-member expert medical team assisted by China arrived in Islamabad, the capital of Pakistan. Pakistani Foreign Minister Qureshi and other government officials were waiting on the tarmac. The supplies brought along with the special plane include 100,000 medical masks, medical protective clothing, and nucleic acid testing reagents.}  \\
&\parbox[c]{11cm}{\textbf{Time}: 2020/3/28 22:46:00}\\
 \hline
\end{tabular}
\end{CJK}
\end{table}

Next, we will further introduce our pretests, the proposed dataset LTCR, and our model \method.

\subsection{Preliminary Test}\label{subsec-pretest}
%% 预实验，证明长文本假新闻的查全比短文本更困难
Based on CHECKED\footnote{https://github.com/cyang03/CHECKED}, we carry out pretests to prove that long-length fake news is harder to be found completely than short news. In the dataset, the lengths of real news mainly range from 20 to 350 characters and those of fake news from 10 to 300. Hence, eighty and one hundred characters are chosen as two borders of long texts and short texts. Splitting the dataset by length of texts, long-length and short-length data are used to evaluate two text classification models, TextRNN~\cite{2016Recurrent} and FastText~\cite{2017Bag}. The results are shown in Table~\ref{pretest}.

\begin{table}[h] 
\caption{Evaluation metrics of TextRNN and FastText on short-text and long-text data. The boundaries are used to divide the dataset into two disjoint parts, long and short texts. Precision and Recall are in terms of fake news. For metrics, the left and right numbers are related to short and long texts, respectively.\label{pretest}}
\begin{tabular}{@{}cccccc@{}}
\toprule
\textbf{Model}	&   \textbf{Boundary}  & \textbf{Accuracy}   & \textbf{Macro F1}   & \textbf{Precision} &   \textbf{Recall}	\\
\midrule
TextRNN &  80 & 0.9351/0.9466 & 0.9156/0.8868 & 0.9161/0.6928 & 1.0/0.9593  \\
TextRNN	& 100 & 0.8723/0.9726 & 0.8531/0.9269 & 0.8286/0.7905 & 1.0/0.9651   \\
FastText & 80 & 0.9027/0.9922 & 0.8679/0.9810 & 0.8792/0.9558 & 1.0/0.9774    \\
FastText & 100 & 0.9043/0.9967 & 0.8931/0.9903 & 0.8693/0.9882 & 0.9943/0.9767 \\
\bottomrule
\end{tabular}
\end{table}

\begin{figure}[h]
\centering
\begin{minipage}[t]{0.49\textwidth}
    \centering
    \includegraphics[width=0.95\linewidth]{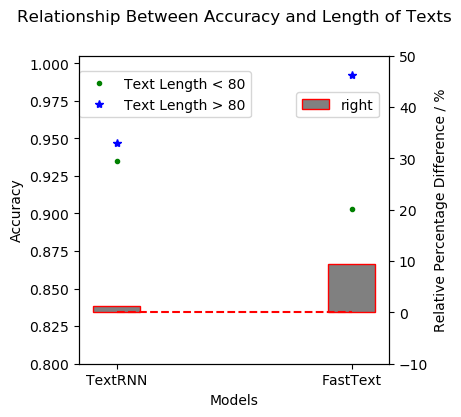}
    \label{fig:sub1}
    \end{minipage}
\begin{minipage}[t]{0.49\textwidth}
    \centering
    \includegraphics[width=0.95\linewidth]{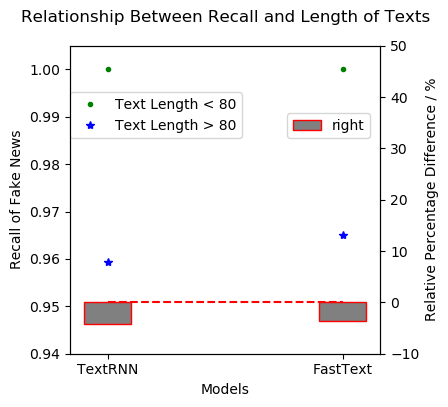}
    \label{fig:sub2}
  \end{minipage}
  \caption{Comparison between the evaluation metrics on short texts and long texts. The fake news recall and accuracy of two models, TextRNN and FastText, on texts shorter and longer than 80 words are scattered in the graph. Relative percentage difference (RPD) between two numbers $a_1$ and $a_2$ is defined as 
$\dfrac{a_2-a_1}{(a_2+a_1)/2}\times 100\%$, representing the difference between the two numbers.}
  \label{fig:pretest-visualization}
%   \end{adjustwidth}
\end{figure}

    The accuracy and macro F1 score of short texts are higher than those of long texts. This is because there are much more real texts than fake ones in the long-text part of the dataset, and real texts are usually easier to be recognized. In those texts longer than a hundred characters, there are 1,652 pieces of real news but only 170 pieces of fake news. 
    Besides, recall, which is the key evaluation metric, of short ones is lower than long ones. To make the conclusion clearer, Figure~\ref{fig:pretest-visualization} presents that for both of the two models, the accuracy on short texts is higher than that on long texts. The recall of short-length fake news exceeds that of the long-length.
    
    Therefore, it is harder to find out all the fake news in long texts than in short texts. High-quality long-text COVID-19 datasets for Chinese fake news detection and better models are required to solve the problem.

\subsection{LTCR Data Collection\label{LTCR}}
% 我们的长文本中文疫情假新闻检测数据集
% Topical subheadings are allowed. Authors must ensure that their Methods section includes adequate experimental and characterization data necessary for others in the field to reproduce their work. Authors are encouraged to include RIIDs where appropriate. 
%% 2. 数据集的构建过程
To achieve good performance on long-text fake news detection, we must make sure that there is as much long-length fake news as possible. However, collecting and annotating raw news on social media are extremely time-consuming and expensive processes. To address this issue, we first accumulate massive amounts of news items from public datasets such as the Chinese COVID-19 Fake News Dataset~\cite{yang2021checked} and DXY-COVID-Rumor~\cite{dxy-rumor}. We also crawl extra fake news items on fact-checking websites. Then, long-length news is selected from these data and cleaned to prevent the noise caused by similar news. The process of our proposed LTCR is detailed below.

\textbf{Investigating available sources of Chinese fake and real news on COVID-19.} Two kinds of sources are considered, which are formal and informal. Formal sources are datasets proposed in published papers. Informal sources include datasets published without authorized scripts and fact-checking websites where rumors are explained. All of the sources are summarized in Table~\ref{dataset-source}. 

\begin{table}[h] 
\caption{Available sources of COVID-19 Chinese fake news and real news.\label{dataset-source}}
\begin{tabular}{@{}ccc@{}}
\toprule
\textbf{Source}	& \textbf{Type}   & \textbf{Time Range}	\\
\midrule
CHECKED 	& 	Formal dataset  & 	December 2019 to August 2020\\
DXY-COVID-Rumor		& Formal dataset	  & January 2020 to March 2020 \\
COVID-19-rumors &   Informal public dataset & January 2020 to April 2020\\
Jiaozhen\footnotemark[1] & Fack-checking website   & January 2020 to January 2023\\
Guangdong News\footnotemark[2] & Fack-checking website   & January 2020 to February 2023\\
Hubei News\footnotemark[3] & Fack-checking website   & February 2020 to April 2022\\
\bottomrule
\end{tabular}

\end{table}

\footnotetext[1]{https://vp.fact.qq.com/home}
\footnotetext[2]{https://news.southcn.com/node\_32f8192c23}
\footnotetext[3]{http://www.cnhubei.com/z/12644102/}

%% 爬取较真辟谣网站的过程，包括keywords的选取（参考CHECKED列一个关键词表）、爬取到的信息（也就是csv中的表头）
\textbf{Crawling rumors on fact-checking websites.} To alleviate the lack of rumors in existing datasets, we collect rumors on fact-checking websites. Rumors on the Guangdong News website and Hubei News website are usually short to help readers get the main idea as soon as possible. For example, a piece of rumor refutal news is posted to clear the rumor saying ``COVID-19 has disappeared''. The contents on the two websites are heterogeneous and not structured, making it hard to extract rumors from the news. For instance, the sentence, ``The rumor is:'', is followed by fake news in a few documents but not in other ones on the websites. This makes it hard to crawl fake news automatically on the two websites. However, on Jiaozhen, a fact-checking website of Tencent, suspicious rumors follow a common sentence, ``It is a popular saying that''. Besides, many sayings are whole stories instead of one short sentence on this website. As a consequence, Jiaozhen is chosen as the main source to be crawled on, while the other two websites are dismissed.

The data is gathered through a Python-based web crawler that scanned all 104 pages. Then we processed the data to extract the ``sayings'' from each article. ``Sayings'' are defined as the primary content of each article that includes the rumors or false news being debunked. Lastly, irrelevant news is removed using a list of COVID-19-related keywords. The choice of keywords follows CHECKED~\cite{yang2021checked}. All of the keywords and their English translations can be found in Table~\ref{key}.

\begin{table}[!ht]
    \centering
    \caption{List of keywords relevant to COVID-19.}
\begin{tabular}{cc}
\hline \text { \textbf{Categories} }  & \text { \textbf{English Translation} } \\
\hline \multirow{2}{*}{$\begin{array}{l}
\text { Coronavirus and COVID-19 }
\end{array}$} 
& \text { Coronavirus, COVID-19} \\
& \text { SARS-CoV-2, COVID}  \\
 \hline
\multirow{2}{*}{\text { Pandemic }} & \text { Pandemic, Epidemic Area, Infection,} \\
& \text {Confirmed Case, Death Case, Imported Case } \\
 \hline
 \multirow{2}{*}{$\begin{array}{l}
\text { Figures and Organizations }
\end{array}$} 
  & \text { WHO, Nanshan Zhong, Wenhong Zhang, }  \\
    & \text { Wenliang Li, Fauci, CDC }  \\
\hline
 \multirow{2}{*}{$\begin{array}{l}
\text { Medical Supplies }
\end{array}$}  
& \text { Testing kit, Nucleic Acid Test, Vaccine, Antibody,} \\
& \text {Huoshenshan, Leishenshan, Mask, N95 } \\
 \hline
 \multirow{2}{*}{\text { Policies }}  & \text { Quarantine, Lockdown, Prevention And Control, Herd Immunity,} \\  &
 \text {Health Code, Combat COVID-19, Love for Wuhan }\\
\hline
\end{tabular}
\label{key}
\end{table}
    
    \textbf{Aggregating data from different sources.} We meet two data conflicts, \textit{Inconsistent Data Headers} and \textit{Similar News Contents}. Similar news contents may introduce noise into the dataset, making the model biased during training. We design methods for solving these conflicts as follows:  % 数据冲突的问题如何解决
    \begin{enumerate}
        \item[(1)] For \textbf{Inconsistent Data Headers}, we select some common headers (i.e., ``id'', ``text'', ``label'' and ``time'') and also some non-shared but probably useful headers (i.e., ``title'' and ``summary''). We use the TextRank4ZH\footnote{https://github.com/letiantian/TextRank4ZH} toolkit to generate titles and summaries for those instances that do not have the two headers.
        \item[(2)] For \textbf{Similar News Contents}, we remove all instances that have similar news content but one. The cosine similarities of each pair of news are calculated by using the Scikit-Learn package \cite{pedregosa2011scikit}. If the similarity between any two texts in a group of texts is greater than 0.8, we remove all but one text from the dataset to improve its quality.
    \end{enumerate}
    In summary, headers for our dataset are ``id'', ``title'', ``summary'', ``text'', ``label'' and ``time'', where ``text'' is the content of the news and ``label'' is zero or one as symbols of fake or real news. Detailed information and explanation can be found in Table~\ref{head}. 
    
    \begin{table}[h]
    \centering
    \caption{Explanation of the headers of our proposed dataset LTCR.}
    \begin{tabular}{cc}
    \hline
        \textbf{Headers} & \textbf{Explanation} \\ \hline
        \textbf{id} & Refers to a unique number that can be used to distinguish between data entries. \\ 
        \textbf{title} & Refers to the title or heading which is used to describe the specific content or theme. \\ 
        \textbf{summary} & Refers to the summary or abstract which provides a brief overview of the content. \\ 
        \textbf{text} & Refers to the main body of text or detailed information. \\ 
        % \textbf{prove} & Refers to the evidence or proof which is used to support the claims or information. \\ 
        \textbf{label} & Refers to the categorization or classification which is used to classify the content. \\ 
        \textbf{time} & Refers to the timestamp or date when the data or information was collected. \\ \hline
    \end{tabular}
    \label{head}
\end{table}

    \textbf{Filtering and cleaning the dataset.} First, the dataset is filtered by the length of news content. Instances with texts shorter than 80 words are deleted to endow our dataset with the feature of long texts. Next, some texts are manually shifted to the form of rumors instead of explanations. It is found that some rumors are not original rumors but descriptions of whether they are fake or real. Hence, the rumors are extracted by hand to enhance the originality of the dataset.
    Besides, Some fake news has obvious characteristics after being crawled. For example, it will be noted at the beginning of the news: There have been rumors recently, etc. We manually remove these irrelevant flags.
    % 这一行是人工数据清洗的工作 
    
\subsection{\method: Salience-Aware Fake News Detection Model}
\begin{figure}[h]
    \centering
    \includegraphics[width=1\textwidth]{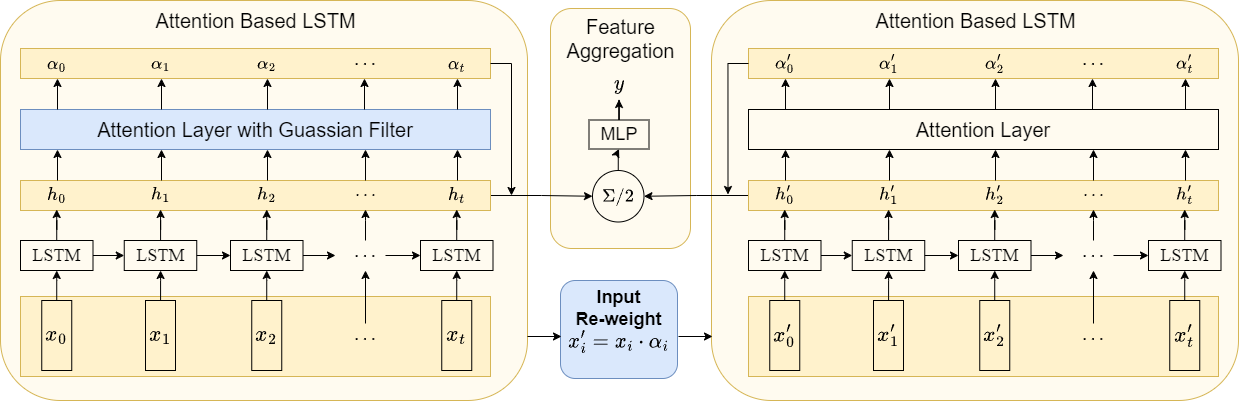}
    \caption{The architecture of the proposed model \method. It consists of two Attention-based LSTM modules and a mechanism called Input Re-weight. $x_i(i=0,1,2,...,t.)$ are text embeddings of every token. $h_i(i=0,1,2,...,t.)$ are outputs of LSTM layers. $\alpha_i(i=0,1,2,...,t.)$ are attention weights, representing the importance of each token.}
    \label{model}

\end{figure}

% 模型
%% 1. 框架图
%% 2. 介绍各个模块，LSTM、Attention、Input Re-weight Module
The proposed model is presented in Figure~\ref{model}. It only has two LSTM (Long Short-Term Memory) layers, enhanced by a new mechanism called Input Re-weight, which is the core of the model. In the field of Computer Vision, a saliency map refers to an image that highlights the regions that attract people's attention at their first glimpse. Similarly, some words or sentences in a text will draw peoples' attention first. These parts usually contain important information and contribute much to the classification. Therefore, we detect the salience of the input in the first LSTM module and pay more attention to those parts that have higher values of salience in the second module.

\textbf{Attention-based LSTM module.} The Attention-based LSTM model is a type of recurrent neural network that is capable of processing sequential data and retaining important information over time. Let's assume an input sequence of length $n$ is represented as $X = {x_1, x_2, ..., x_n}$. The output of the LSTM layer is represented as $H = {h_1, h_2, ..., h_n}$, where $h_i$ is the hidden state of the LSTM at time step $i$. Then, the attention weights $\alpha = {\alpha_1, \alpha_2, ..., \alpha_n}$ are computed. These weights represent the importance of each element in the input sequence. The attention weights are computed using a separate neural network that takes the LSTM output as input:

\begin{equation}
    e_i = v^T * tanh(W_h * h_i + b); \quad \alpha_i = softmax(e_i)
\end{equation}

In these equations, $W_h$ and $b$ are learnable parameters of the attention mechanism, $v$ is a learnable parameter used to compute the scalar product with the output of the attention mechanism, and softmax is a function that normalizes the attention weights, to sum up to one. The Attention mechanism in this model allows the network to focus on specific parts of the input sequence when making predictions, rather than treating all parts of the sequence equally. This helps to improve the accuracy of the model by allowing it to give more weight to the most relevant parts of the input sequence.

The context vector $c$ is a weighted sum of the LSTM output, which is the final output of the module.
\begin{equation}
    c = sum(\alpha_i * h_i)
\end{equation}

Additionally, to enlarge the range of salient texts that get more attention, we add a Gaussian Filter after the regular Attention Layer. Gaussian filter has the ability to smooth a distribution. After being filtered, the numbers that are close in terms of positions become closer in terms of their values. This means that filtering the attention weights using a Gaussian Filter will make the context of the tokens with high weights more important.
\begin{equation}
    \mathbf{\alpha} = Gaussian\_filter(\mathbf{\alpha})
\end{equation}

\textbf{Input Re-weight module.} To tackle the problem that it is hard to find all the fake news when their lengths get longer, the Input Re-weight module is proposed. When people judge a piece of news, they usually go through the whole news to find suspicious parts which are possible to be fake and then read it repeatedly to check its veracity. Inspired by this, we utilized Attention Mechanism to reweigh all the tokens of a text, intimating the process of locating important sentences. 
\begin{equation}
    x_i' = x_i * \alpha_i
\end{equation}

\textbf{Feature aggregation.} Passing the reweighed input to a new Attention-based LSTM module, the model checks the news one more time. Finally, the context vectors of the two LSTM layers are averaged as the final context vector. Passing it to an MLP (Multi-Layer Perceptron), the model outputs the probabilities of each class.
\begin{equation}
    h_i' = LSTM(x_i'); \quad
    \alpha_i' = Attention(h_i');  \quad
    c' = sum(\alpha_i' * h_i')  
\end{equation}

\begin{equation}
    c_{final}  = \dfrac{c+c'}{2};   \quad
    y = MLP(c_{final})
\end{equation}

\section{Data Analysis}\label{sec4}
% Corpus Statistics
\subsection{Dataset Statistics}
In this section, we present an analysis of our LTCR dataset, which contains news articles from different sources. The dataset is cleaned and processed to extract sayings from each article, and instances with texts shorter than 80 words are removed.

\begin{figure}[h]
\centering
\includegraphics[width=0.6\textwidth]{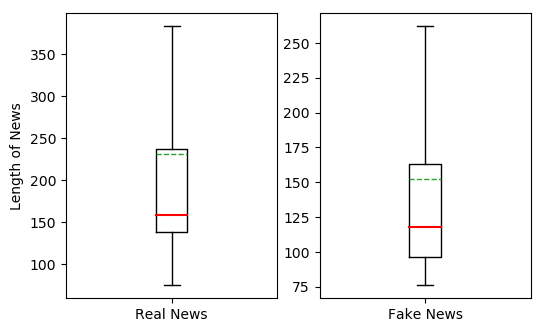}
\caption{The box plot of text lengths on the LTCR dataset. The red lines are the medians. The green lines represent the average numbers.}
\label{fig:dataset}
\end{figure}

Table~\ref{stat} reports the data statistics such as the number of news,Maximum length and average length, 
The dataset contains a total of 2,290 news articles, with 1,729 (75.5\%) labeled as real news and 561 (24.5\%) labeled as fake news. The distribution of text length are visualized in Figure~\ref{fig:dataset}.

\begin{table}[!ht]
    \centering
    \caption{Descriptive statistics of text length (in words) for real and fake news articles in the LTCR dataset.}
    \begin{tabular}{ccccc}
    \hline
        \textbf{} & \textbf{number of news} &\textbf{Maximum length} & \textbf{average length} & \textbf{Standard Deviation} \\ \hline
        \textbf{Real }& 1,729 & 4,984 & 232.5 & 256.6 \\ 
        \textbf{Fake} & 561 & 2,309 & 153.5 & 133.1 \\ \hline
    \end{tabular}
    \label{stat}
\end{table}

\begin{figure}[h]
    \centering
    \includegraphics[width=\textwidth=1.3]{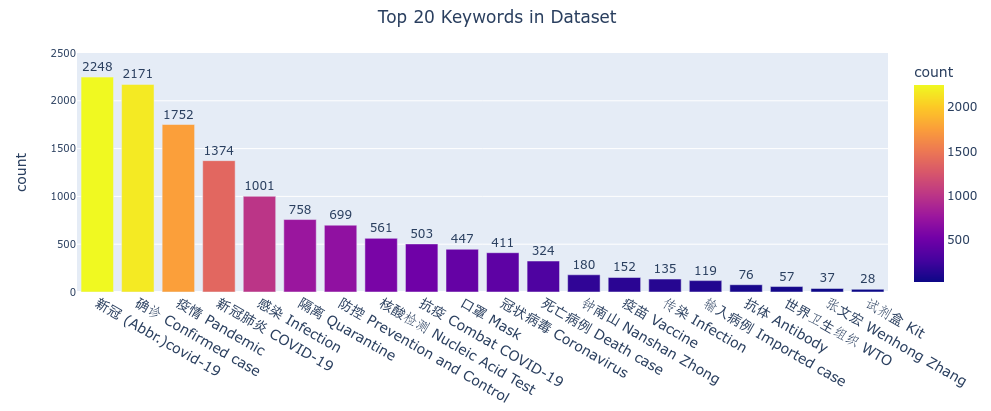}
    \caption{Distribution of keywords.}
    \label{fig:key}
\end{figure}

Meanwhile, in order to ensure the comprehensiveness of the data, we collected news on various topics to form our dataset and we classified the news using keywords. The distribution of keywords for the entire dataset is shown in Figure~\ref{fig:key}.

\subsection{Data Comparison}

We select two authoritative Chinese rumor detection datasets, CHECKED and DXY-COVID-Rumor, to compare with the dataset we construct. We will compare our dataset from the following dimensions to demonstrate its superiority: number of news, maximum length of news, minimum length of news, average length of news and median length of news. The comparison results are shown in Table \ref{compared}.

\begin{figure}[h]
    \centering
    \includegraphics[width=\textwidth=1.2]{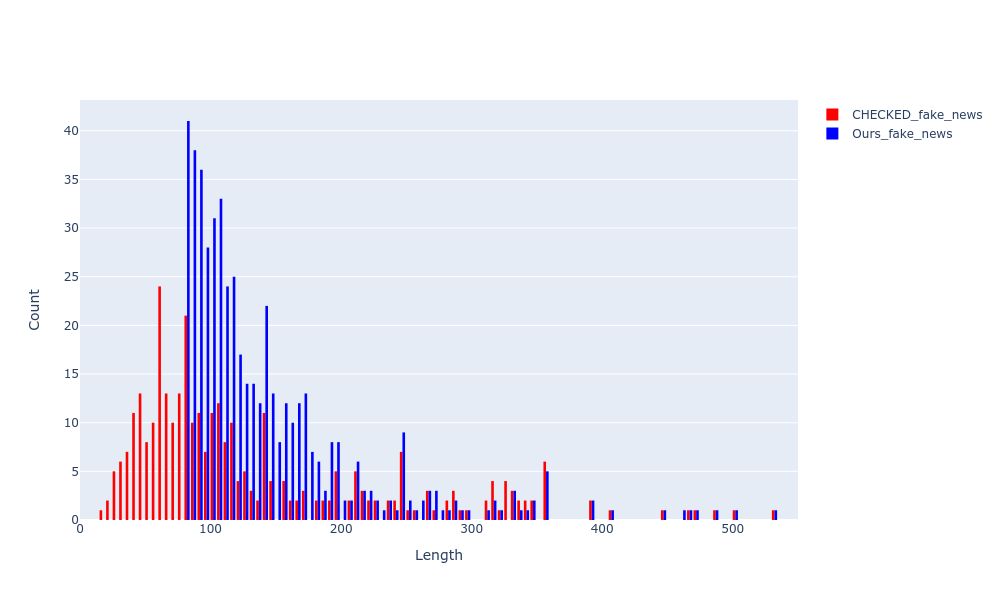}
    \caption{The comparison of the length of fake news with the CHECKED dataset.}
    \label{fake2}
\end{figure}

At the same time, to better illustrate the length distribution of fake news in our dataset, we compared it with the fake news in the CHECKED dataset. The horizontal axis represents the length of the news, and the vertical axis represents the number of news articles corresponding to that length. The results are shown in Figure \ref{fake2}.

\begin{table}[!ht]
    \centering
    \caption{Comparison on statistics of our dataset and the other two formal datasets.}
    \begin{tabular}{c|c|cccc}
    \hline
    \multicolumn{2}{c|}{Dataset}
           & Maximum length & Minimum length & Average length & median length\\ \hline
        Ours & real  & 4,984 & 80 & 232.5 &161 \\ 
        ~ & fake  & 2,309 & 80 & 155.5  &120\\ \hline
        CHECKED\cite{yang2021checked} & real  & 3,778 & 13 & 228.5 & 159.0  \\ 
        ~ & fake & 2,267 & 11 & 150.8 & 99.5 \\ \hline
        DXY-COVID-Rumor\cite{dxy-rumor} & real  & 3,658 & 8 & 260.6 & 200.0 \\ 
        ~ & fake & 180 & 15 & 112.4 & 109.7 \\ \hline
    \end{tabular}
    \label{compared}
\end{table}

Due to the fact that long text genuine news is very easy to find, but most rumors on the internet are short sentences rather than lengthy speeches, the number of long text fake news is very small. Therefore, our dataset focuses on long text fake news rather than real news. Through comparison, we can find that our dataset has the highest number of long text fake news, while the average length and minimum length are higher than the other two datasets. This indicates that the news we construct is composed of long text news, which has a clear advantage in news length. 

In order to show the differences between the three datasets more clearly, we listed their longest news and shortest news in Table \ref{sample} from the fake news of the three datasets.

\begin{table}[!ht]
    \centering
    \caption{The shortest fake news from the three datasets.}
    \begin{tabular}{c|l}
    \hline
        ~ & \textbf{The shortest fake news:}{\begin{CJK}{UTF8}{gbsn} 美国瑞德西韦已获川普总统特批......自己生产这个药。（共80字）\end{CJK}}\\ 
             Ours & \textbf{Translation:} American remdesivir has been specially approved by President Trump......\\
               ~ & \quad \quad \quad \quad \quad \quad \quad Chinese companies can produce this medicine themselves (80 words in total) \\ 
        % ~ & \textbf{Longest fake news:}{\begin{CJK}{UTF8}{gbsn}俄罗斯确诊超过一万，一怒之下驱赶了上百万中国人......\end{CJK}}\\
        %  ~ & \quad \quad \quad \quad \quad \quad \quad \quad \quad \quad {\begin{CJK}{UTF8}{gbsn}我们更应该指出其不当之处。（共2309字）\end{CJK}} \\ 
        %    ~ & \textbf{Translation:}Russia has been diagnosed with over 10000 cases, driving away \\
        %    ~ & millions of Chinese people in a fit of anger We should also point out its shortcomings. (2309 words in total) \\    
       \hline

        ~ & \textbf{The shortest fake news:} {\begin{CJK}{UTF8}{gbsn} 周杰伦去福州自备隔离仓。 \end{CJK}}\\ 
          CHECKED & \textbf{Translation:}Jay Chou brought his own quarantine warehouse when he went to Fuzhou. \\ 
        % ~ & \textbf{Longest fake news:}{\begin{CJK}{UTF8}{gbsn}目前武汉肺炎已经是三代或四代感染......南昌大学第一附属医院。（共2267字）\end{CJK}} \\ 
        %   ~ & \textbf{Translation:} \\ 
        \hline

        ~ & \textbf{The shortest fake news:} {\begin{CJK}{UTF8}{gbsn}降压药会增加感染新冠病毒的风险。\end{CJK}} \\ 
           DXY-COVID-Rumor & \textbf{Translation:} Antihypertensive drugs increase the risk of infection with COVID-19.\\

          %   ~ & \textbf{Longest fake news:}{\begin{CJK}{UTF8}{gbsn}如果喉咙湿润的话，病毒大多会沿食道进入胃部……对肾脏等器官造成额外负担。(共180字)\end{CJK}} \\ 
          % ~ & \textbf{Translation:} \\ 
        \hline

    \end{tabular}
    \label{sample}
\end{table}

The above analysis indicates that: 
\begin{enumerate}
    \item[(1)] Compared with other Chinese rumor datasets, the LTCR dataset focuses on long-text data, contains longer and more usable fake news texts, filling the gap in Chinese long-text rumor detection datasets related to COVID-19.
    \item[(2)] The LTCR dataset is larger and more representative than existing Chinese fake news datasets. The sources of the dataset include multiple Chinese news websites, containing many high-quality true and false news articles.
    \item[(3)] The LTCR dataset is more comprehensive and covers multiple topics of the epidemic, which can improve the generalization performance of the model, inform future research on detecting rumors and fake news related to COVID-19.
    \item[(4)] The news articles selected for our dataset have all been spread on the internet. We have carefully screened these articles for their logical and confusing nature, making them suitable for research and application in Chinese long-text fake news detection. This can help researchers better understand the characteristics of Chinese fake news and improve the accuracy and efficiency of Chinese long-text fake news detection.
\end{enumerate}

\section{Experiments}\label{sec5}
\subsection{Evaluation Metrics}
% 评价指标：准确率、查准率、查全率、F-score
To quantitatively evaluate our method, we adopt Accuracy, Precision, Recall, and F-Score as the evaluation metrics.

We use F1 Score to evaluate the performance of our model to identify fake news (samples with the negative label), which is important in the scenario of fake news detection. F-Score is the harmonic average of Recall and Precision. Recall
quantifies the ability to detect a piece of rumor when it is actually fake, and Precision is the proportion of predicted fake news that is actually fake. Because missing fake news is more unaffordable than making inaccurate classification, we increase the weight of Recall in the formulation of the F-score. The formulas can be given by
\begin{equation}
    Recall = \dfrac{TP}{TP+FN},
\end{equation}
\begin{equation}
    Precision = \dfrac{TP}{TP+FP},
\end{equation}
\begin{equation}
    F-score = \dfrac{(1+\beta)Precision*Recall}{\beta*Precision+Recall},\quad \beta=2,
\end{equation}
where TP, FN, and FP indicate true positives, false negatives, and false positives, respectively.
\subsection{Baseline Methods and Ablation Approaches}
% 与 Baselines 对比(在我们自己的数据集上实验)
 We evaluate \method\ and baselines in terms of Accuracy, Recall, Precision, and F-score on the LTCR dataset. The baselines include models that have achieved significant results in the field of text classification in recent years,e.g.,BertGCN,InducT-GCN.
 and some classic text classification models, e.g., TextCNN ~\cite{2014Convolutional}, TextRNN ~\cite{2016Recurrent}.
 
%% 介绍所有用到的Baselines: TextCNN, TextRNN, TextRCNN, TextRNN+Attn, Transformer
\begin{enumerate}

    \item[(1)] \textit{ InducT-GCN}~\cite{wang2022inductgcn} It uses Inductive Graph Convolutional Networks to classify text by treating it as a graph and capturing semantic relationships between words. 

     \item[(2)] \textit{BertGCN}~\cite{2021BertGCN} It combines BERT and GCN to extract high-quality features from text and capture interdependencies between words. This approach can effectively understand semantic information and classify the text.

    \item[(3)] \textit{DeBERTa (Decoding-enhanced BERT with disentangled attention)}~\cite{he2021deberta} enhances the BERT model by disentangling content and position information in self-attention. It uses a relative position bias to improve the model's performance in text classification tasks.

   \item[(4)] \textit{ ALBERT (A Lite BERT) }~\cite{he2021deberta} It is a lighter version of BERT for text classification. ALBERT uses factorized embedding parameterization and cross-layer parameter sharing to reduce the model size while maintaining performance.
   
  \item[(5)] \textit{FastText}~\cite{2017Bag} It represents text as a bag of words and uses n-gram features. It is effective for rumor detection as it captures important text features and classifies text accordingly.

    \item[(6)] \textit{DPCNN}~\cite{2017Deep} It extracts features using convolutional and pooling layers with shortcut connections to preserve information and create a deep pyramid structure. For rumor detection, DPCNN can extract hierarchical features to detect rumors in text.

    \item[(7)] \textit{Transformer}~\cite{2017attention} It is a neural network that captures word dependencies with self-attention and processes text in parallel with multi-head attention. It can detect rumors by capturing these dependencies and aspects of the text.

    \item[(8)] \textit{TextRNN+attn}~\cite{2016attention} It is a recurrent neural network that captures sequential information and important text features with attention. This approach effectively captures important text features and can classify text to detect rumors.

    \item[(9)] \textit{TextCNN}~\cite{2014Convolutional} It extracts features by convolving filters over word embeddings, capturing local patterns and pooling them to obtain a fixed-length text representation. This representation is classified to detect rumors based on extracted features.

    \item[(10)] \textit{TextRCNN}~\cite{2015Recurrent} It processes text sequentially to capture sequence information. The model uses a hidden state to update and capture the context of the text, and can effectively detect rumors by leveraging this sequence information.
   
    \item[(11)] \textit{w/o Salience} To analyze the effectiveness of the proposed Input Re-weight module, we conduct an ablation test, removing the module and setting the input of the second LSTM module as the original text embeddings.
\end{enumerate}

\subsection{Implementation Details}
% 实验细节
We use PyTorch version 1.1 to implement the model mentioned earlier. Data is inputted into the model on a per-character basis, and pre-trained word vectors from Sogou News' Word \& Character 300d version \cite{P18-2023} are utilized. It is simple to acquire pre-trained vectors with varying attributes and apply them to downstream tasks.

In our experiments, we divided the training set, testing set, and validation set in a 3:1:1 ratio. We employ mini-batch gradient descent to train our text classification model. We set the batch size to 128 and each sequence length to 256. For each epoch, we iterate through the entire dataset 20 times and use an optimizer with a learning rate of 0.001 to update the model's weights. To mitigate the risk of overfitting, we utilize a 50\% dropout rate, meaning that each neuron had a 50\% chance of being dropped during forward propagation. If the performance does not improve for 1,000 consecutive batches, we terminate the training early to save time and computing resources.

\subsection{Experimental Results}

%% 实验结果
The evaluation metrics measured on the testing data are demonstrated in Table~\ref{experiments}. Our model outperforms other baselines in terms of Recall and F-score, proving that the \method\ has a stronger ability to fully detect fake news. We observe the following:
\begin{enumerate}
    \item[(1)]  The recall of our model (90.91\%) ranks first compared with other baselines. The results show that, by analyzing different parts of a text with matched levels of salience, the model gains a deeper understanding of the texts and will not regard a hard-to-tell text as real arbitrarily. 
    
    \item[(2)]  Most baseline models have high precision and accuracy but low recall. This is probably because they tend to classify uncertain news as real and are unable to discriminate the news by looking at details. 
    
    \item[(3)]  Attention Mechanism can effectively enhance RNN-based models. TextRNN gets a full score on the recall of real news as it predicts all the news as real news. After adding Attention Mechanism, TextRNN performs much better with higher accuracy (93.89\%), recall (79.8\%), precision (90.8\%), and F-score (83.16\%). 
    
    \item[(4)]  The ablation test for the salience-aware Input Re-weight module shows that the improvement of \method\ comes from the Re-weight module instead of the increase of layers. By comparing the model w/o Re-weight with the Attention-based LSTM model and \method, we find that the recall of the model with two paralleled LSTM layers is roughly three percentage points higher than that of the one-layer model TextRNN+Attn. After adding the Re-weight module, the recall increases by approximately nine percentage points and the F-score rises from 86.01\% to 89.51\%, which represents a significant improvement. Therefore, the ability of the \method\ mainly comes from the Re-weight module. The intuition is that reading a text with attention for the second time is more useful than reading two times without any focus.  
\end{enumerate}

\begin{table}[h] 
\caption{Evaluation metrics of baselines and the proposed model on Long-Text Dataset for Chinese Fake News Detection.\label{experiments}}
\begin{tabular}{@{}cccccc@{}}\toprule
\textbf{Model}	&   \textbf{Accuracy} & \textbf{Recall}   & \textbf{Precision} &  \textbf{F-score}\\
\midrule

TextRCNN (Lai et al., 2015) & 0.9432 & 0.8182 & 0.9101 & 0.8467    \\

TextCNN (Chen, 2015)  & 0.9389 & 0.8182 & 0.8901 & 0.8408 \\

TextRNN+Attn (Peng et al., 2016)\footnotemark[1] & 0.9389 & 0.7980 & 0.9080 & 0.8316 \\
Transformer (Vaswani et al., 2017) & 0.9410 & 0.8687 & 0.8609 & 0.8661 \\

DPCNN (Johnson et al., 2017) & 0.9541 & 0.8788 & 0.9062 & 0.8877 \\

FastText (Joulin and Tong, 2017) & 0.9345 & 0.7879 & 0.8966 & 0.8211 \\

ALBERTv2 (Lan et al., 2020)  & 0.8983 &\underline{0.8983} & 0.8697   & 0.8837   \\

DeBERTa (He et al., 2021)  & \underline{0.9566} & 0.8961    & \underline{0.9212}    & \underline{0.8973 }   \\

BertGCN (Lin et al., 2021)  & 0.9394 & 0.6338 & 0.5986 & 0.6149 \\

InductTGCN (Wang et al., 2022)  & 0.7837 & 0.7800 & 0.7550 & 0.7596 \\

\method\  (Ours) & \textbf{0.9585} & \textbf{0.9091} & 0.9000 & \textbf{0.9060} \\
\hline
w/o Salience & 0.9498 & 0.8283 & \textbf{0.9318}  & 0.8601\\
\bottomrule
\end{tabular}

\footnotetext[1]{Attn stands for Attention.}

\end{table}

\section{Discussion}\label{sec6}
% 补充说明本文方法的优点（与前面实验证明的优点不同，此处可以写不是那么重要、但是确实存在的优点）
We proposed the long-text Chinese detection dataset (LTCR) and the salience-aware fake news detection model (\method) for healthcare fake news detection. The dataset contains 1,729 and 500 pieces of real and fake news that are all longer than eighty characters. It provides the models with long-length training data to enhance their ability to detect long-text fake news. The evaluation is more expressive than the previous benchmark CHECKED because LTCR contains more fake news.

The design of the \method is quite simple with only two LSTM layers. The Input Re-weight module is a general idea that can be plugged into other models to extract more detailed features. Different from cascading layers tail-to-head, reweighing inputs means referencing the original inputs with attention instead of continuing detection based on features in the middle. However, there are some drawbacks to the model and experiments. We do not explore the competence of models with more layers. It is unsure in which layer the improvement brought by the Input Re-weight module becomes trivial.

% Discussions should be brief and focused. In some disciplines use of Discussion or `Conclusion' is interchangeable. It is not mandatory to use both. Some journals prefer a section `Results and Discussion' followed by a section `Conclusion'. Please refer to Journal-level guidance for any specific requirements. 

\section{Conclusions and Future Work}\label{sec7}
COVID-19 fake news on social media impacts the behaviors and mindset of patients and medical workers. This study finds that long texts are more difficult to be detected completely than short ones. To tackle the problem that the lack of fake news, especially long-length fake news, in existing datasets leads to poor ability to detect long-text rumors, a long-length COVID-19 Chinese rumor detection dataset LTCR is proposed. Moreover, \method, a Salience-Aware Fake News Detection model, shows the best performance in finding as much fake news as possible according to extensive experimental results. We hope that our work will help improve long-text fake news detection, especially on Chinese social media such as Weibo.

% future research directions
The model \method\ in the paper is inspired by the working process of human beings. Continuing with this, three future research directions are considered. For starters, MultiCheck, the model with more than two layers of LSTM connected by the Input Re-weight module, is on the list. This studies the upper limit of improvement that the module can bring to text classification. The second one is to apply the idea of reweighing inputs according to attention weights to typical text classification tasks. Such kind of models cost much fewer computation resources and less time than large pre-trained language models. The third direction is to dig for other ideas from human-being behaviors. These ideas are more explainable to people and usually effective in improving existing models.
% Conclusions may be used to restate your hypothesis or research question, restate your major findings, explain the relevance and the added value of your work, highlight any limitations of your study, describe future directions for research and recommendations. 

% In some disciplines use of Discussion or 'Conclusion' is interchangeable. It is not mandatory to use both. Please refer to Journal-level guidance for any specific requirements. 

\backmatter

% If your article has accompanying supplementary file/s please state so here. 

% Authors reporting data from electrophoretic gels and blots should supply the full unprocessed scans for key as part of their Supplementary information. This may be requested by the editorial team/s if it is missing.

% Please refer to Journal-level guidance for any specific requirements.

% Acknowledgments are not compulsory. Where included they should be brief. Grant or contribution numbers may be acknowledged.

% Please refer to Journal-level guidance for any specific requirements.

%%%%%%%%%%%%%%%%%%%%%%%%%%%%%%%%%%%%%%%%%%

%\bibliographystyle{plainnat}
\bibliography{sn-bibliography}% common bib file


\begin{thebibliography}{38}
\ifx \bisbn   \undefined \def \bisbn  #1{ISBN #1}\fi
\ifx \binits  \undefined \def \binits#1{#1}\fi
\ifx \bauthor  \undefined \def \bauthor#1{#1}\fi
\ifx \batitle  \undefined \def \batitle#1{#1}\fi
\ifx \bjtitle  \undefined \def \bjtitle#1{#1}\fi
\ifx \bvolume  \undefined \def \bvolume#1{\textbf{#1}}\fi
\ifx \byear  \undefined \def \byear#1{#1}\fi
\ifx \bissue  \undefined \def \bissue#1{#1}\fi
\ifx \bfpage  \undefined \def \bfpage#1{#1}\fi
\ifx \blpage  \undefined \def \blpage #1{#1}\fi
\ifx \burl  \undefined \def \burl#1{\textsf{#1}}\fi
\ifx \doiurl  \undefined \def \doiurl#1{\url{https://doi.org/#1}}\fi
\ifx \betal  \undefined \def \betal{\textit{et al.}}\fi
\ifx \binstitute  \undefined \def \binstitute#1{#1}\fi
\ifx \binstitutionaled  \undefined \def \binstitutionaled#1{#1}\fi
\ifx \bctitle  \undefined \def \bctitle#1{#1}\fi
\ifx \beditor  \undefined \def \beditor#1{#1}\fi
\ifx \bpublisher  \undefined \def \bpublisher#1{#1}\fi
\ifx \bbtitle  \undefined \def \bbtitle#1{#1}\fi
\ifx \bedition  \undefined \def \bedition#1{#1}\fi
\ifx \bseriesno  \undefined \def \bseriesno#1{#1}\fi
\ifx \blocation  \undefined \def \blocation#1{#1}\fi
\ifx \bsertitle  \undefined \def \bsertitle#1{#1}\fi
\ifx \bsnm \undefined \def \bsnm#1{#1}\fi
\ifx \bsuffix \undefined \def \bsuffix#1{#1}\fi
\ifx \bparticle \undefined \def \bparticle#1{#1}\fi
\ifx \barticle \undefined \def \barticle#1{#1}\fi
\bibcommenthead
\ifx \bconfdate \undefined \def \bconfdate #1{#1}\fi
\ifx \botherref \undefined \def \botherref #1{#1}\fi
\ifx \url \undefined \def \url#1{\textsf{#1}}\fi
\ifx \bchapter \undefined \def \bchapter#1{#1}\fi
\ifx \bbook \undefined \def \bbook#1{#1}\fi
\ifx \bcomment \undefined \def \bcomment#1{#1}\fi
\ifx \oauthor \undefined \def \oauthor#1{#1}\fi
\ifx \citeauthoryear \undefined \def \citeauthoryear#1{#1}\fi
\ifx \endbibitem  \undefined \def \endbibitem {}\fi
\ifx \bconflocation  \undefined \def \bconflocation#1{#1}\fi
\ifx \arxivurl  \undefined \def \arxivurl#1{\textsf{#1}}\fi
\csname PreBibitemsHook\endcsname

%%% 1
\bibitem[\protect\citeauthoryear{El-Far~Cardo et~al.}{2021}]{intro-impacts}
\begin{barticle}
\bauthor{\bsnm{El-Far~Cardo}, \binits{A.}},
\bauthor{\bsnm{Kraus}, \binits{T.}},
\bauthor{\bsnm{Kaifie}, \binits{A.}}:
\batitle{Factors that shape people's attitudes towards the covid-19 pandemic in
  germany-the influence of media, politics and personal characteristics}.
\bjtitle{International journal of environmental research and public health}
\bvolume{18}(\bissue{15}),
\bfpage{7772}
(\byear{2021})
\doiurl{10.3390/ijerph18157772}
\end{barticle}
\endbibitem

%%% 2
\bibitem[\protect\citeauthoryear{van~der Linden
  et~al.}{2020}]{intro-fake-cures}
\begin{barticle}
\bauthor{\bsnm{Linden}, \binits{S.}},
\bauthor{\bsnm{Roozenbeek}, \binits{J.}},
\bauthor{\bsnm{Compton}, \binits{J.}}:
\batitle{Inoculating against fake news about covid-19}.
\bjtitle{Frontiers in psychology}
\bvolume{11},
\bfpage{566790}
(\byear{2020})
\doiurl{10.3389/fpsyg.2020.566790}
\end{barticle}
\endbibitem

%%% 3
\bibitem[\protect\citeauthoryear{Weng and Lin}{2022}]{intro-fake-conspiracy}
\begin{barticle}
\bauthor{\bsnm{Weng}, \binits{Z.}},
\bauthor{\bsnm{Lin}, \binits{A.}}:
\batitle{Public opinion manipulation on social media: Social network analysis
  of twitter bots during the covid-19 pandemic}.
\bjtitle{International journal of environmental research and public health}
\bvolume{19}(\bissue{24}),
\bfpage{16376}
(\byear{2022})
\doiurl{10.3390/ijerph192416376}
\end{barticle}
\endbibitem

%%% 4
\bibitem[\protect\citeauthoryear{Mousoulidou
  et~al.}{2023}]{intro-dna-conspiracy}
\begin{barticle}
\bauthor{\bsnm{Mousoulidou}, \binits{M.}},
\bauthor{\bsnm{Christodoulou}, \binits{A.}},
\bauthor{\bsnm{Siakalli}, \binits{M.}},
\bauthor{\bsnm{Argyrides}, \binits{M.}}:
\batitle{The role of conspiracy theories, perceived risk, and trust in science
  on covid-19 vaccination decisiveness: Evidence from cyprus}.
\bjtitle{International Journal of Environmental Research and Public Health}
\bvolume{20}(\bissue{4}),
\bfpage{2898}
(\byear{2023})
\doiurl{10.3390/ijerph20042898}
\end{barticle}
\endbibitem

%%% 5
\bibitem[\protect\citeauthoryear{Ravichandran and
  Keikhosrokiani}{2023}]{ravichandran2023classification}
\begin{barticle}
\bauthor{\bsnm{Ravichandran}, \binits{B.D.}},
\bauthor{\bsnm{Keikhosrokiani}, \binits{P.}}:
\batitle{Classification of covid-19 misinformation on social media based on
  neuro-fuzzy and neural network: A systematic review}.
\bjtitle{Neural Computing and Applications}
\bvolume{35}(\bissue{1}),
\bfpage{699}--\blpage{717}
(\byear{2023})
\end{barticle}
\endbibitem

%%% 6
\bibitem[\protect\citeauthoryear{Cui and Lee}{2020}]{cui2020coaid}
\begin{botherref}
\oauthor{\bsnm{Cui}, \binits{L.}},
\oauthor{\bsnm{Lee}, \binits{D.}}:
Coaid: Covid-19 healthcare misinformation dataset.
arXiv preprint arXiv:2006.00885
(2020)
\end{botherref}
\endbibitem

%%% 7
\bibitem[\protect\citeauthoryear{Cheng et~al.}{2021}]{cheng2021covid}
\begin{barticle}
\bauthor{\bsnm{Cheng}, \binits{M.}},
\bauthor{\bsnm{Wang}, \binits{S.}},
\bauthor{\bsnm{Yan}, \binits{X.}},
\bauthor{\bsnm{Yang}, \binits{T.}},
\bauthor{\bsnm{Wang}, \binits{W.}},
\bauthor{\bsnm{Huang}, \binits{Z.}},
\bauthor{\bsnm{Xiao}, \binits{X.}},
\bauthor{\bsnm{Nazarian}, \binits{S.}},
\bauthor{\bsnm{Bogdan}, \binits{P.}}:
\batitle{A covid-19 rumor dataset}.
\bjtitle{Frontiers in Psychology}
\bvolume{12},
\bfpage{644801}
(\byear{2021})
\end{barticle}
\endbibitem

%%% 8
\bibitem[\protect\citeauthoryear{Hossain et~al.}{2020}]{hossain2020covidlies}
\begin{bchapter}
\bauthor{\bsnm{Hossain}, \binits{T.}},
\bauthor{\bsnm{Logan~IV}, \binits{R.L.}},
\bauthor{\bsnm{Ugarte}, \binits{A.}},
\bauthor{\bsnm{Matsubara}, \binits{Y.}},
\bauthor{\bsnm{Young}, \binits{S.}},
\bauthor{\bsnm{Singh}, \binits{S.}}:
\bctitle{Covidlies: Detecting covid-19 misinformation on social media}.
In: \bbtitle{Proceedings of the 1st Workshop on NLP for COVID-19 (Part 2) at
  EMNLP 2020}
(\byear{2020})
\end{bchapter}
\endbibitem

%%% 9
\bibitem[\protect\citeauthoryear{Yang et~al.}{2021}]{yang2021checked}
\begin{barticle}
\bauthor{\bsnm{Yang}, \binits{C.}},
\bauthor{\bsnm{Zhou}, \binits{X.}},
\bauthor{\bsnm{Zafarani}, \binits{R.}}:
\batitle{Checked: Chinese covid-19 fake news dataset}.
\bjtitle{Social Network Analysis and Mining}
\bvolume{11}(\bissue{1}),
\bfpage{58}
(\byear{2021})
\end{barticle}
\endbibitem

%%% 10
\bibitem[\protect\citeauthoryear{Gharaibeh et~al.}{2022}]{related-work-survey}
\begin{botherref}
\oauthor{\bsnm{Gharaibeh}, \binits{M.}},
\oauthor{\bsnm{Obeidat}, \binits{R.}},
\oauthor{\bsnm{Abdullah}, \binits{M.}},
\oauthor{\bsnm{Al-Harahsheh}, \binits{Y.}}:
Datasets and Approaches of COVID-19 Misinformation Detection: A Survey.
IEEE
(2022).
\doiurl{10.1109/icics55353.2022.9811124}
\end{botherref}
\endbibitem

%%% 11
\bibitem[\protect\citeauthoryear{Micallef et~al.}{2020}]{micallef2020role}
\begin{bchapter}
\bauthor{\bsnm{Micallef}, \binits{N.}},
\bauthor{\bsnm{He}, \binits{B.}},
\bauthor{\bsnm{Kumar}, \binits{S.}},
\bauthor{\bsnm{Ahamad}, \binits{M.}},
\bauthor{\bsnm{Memon}, \binits{N.}}:
\bctitle{The role of the crowd in countering misinformation: A case study of
  the covid-19 infodemic}.
In: \bbtitle{2020 IEEE International Conference on Big Data (big Data)},
pp. \bfpage{748}--\blpage{757}
(\byear{2020}).
\bcomment{IEEE}
\end{bchapter}
\endbibitem

%%% 12
\bibitem[\protect\citeauthoryear{Gupta et~al.}{2012}]{2012Evaluating}
\begin{bchapter}
\bauthor{\bsnm{Gupta}, \binits{M.}},
\bauthor{\bsnm{Zhao}, \binits{P.}},
\bauthor{\bsnm{Han}, \binits{J.}}:
\bctitle{Evaluating event credibility on twitter}.
In: \bbtitle{SIAM International Conference on Data Mining}
(\byear{2012})
\end{bchapter}
\endbibitem

%%% 13
\bibitem[\protect\citeauthoryear{Huang et~al.}{2020}]{2020Deep}
\begin{botherref}
\oauthor{\bsnm{Huang}, \binits{Q.}},
\oauthor{\bsnm{Zhou}, \binits{C.}},
\oauthor{\bsnm{Wu}, \binits{J.}},
\oauthor{\bsnm{Liu}, \binits{L.}},
\oauthor{\bsnm{Wang}, \binits{B.}}:
Deep spatial–temporal structure learning for rumor detection on twitter.
Neural Computing and Applications
(3)
(2020)
\end{botherref}
\endbibitem

%%% 14
\bibitem[\protect\citeauthoryear{Trivedi et~al.}{2018}]{2018LinkNBed}
\begin{botherref}
\oauthor{\bsnm{Trivedi}, \binits{R.}},
\oauthor{\bsnm{Sisman}, \binits{B.}},
\oauthor{\bsnm{Dong}, \binits{X.L.}},
\oauthor{\bsnm{Faloutsos}, \binits{C.}},
\oauthor{\bsnm{Ma}, \binits{J.}},
\oauthor{\bsnm{Zha}, \binits{H.}}:
LinkNBed: Multi-Graph Representation Learning with Entity Linkage.
Association for Computational Linguistics
(2018)
\end{botherref}
\endbibitem

%%% 15
\bibitem[\protect\citeauthoryear{Pérez-Rosas et~al.}{2017}]{V2017Automatic}
\begin{botherref}
\oauthor{\bsnm{Pérez-Rosas}, \binits{V.}},
\oauthor{\bsnm{Kleinberg}, \binits{B.}},
\oauthor{\bsnm{Lefevre}, \binits{A.}},
\oauthor{\bsnm{Mihalcea}, \binits{R.}}:
Automatic detection of fake news
(2017)
\end{botherref}
\endbibitem

%%% 16
\bibitem[\protect\citeauthoryear{Zhou et~al.}{2019}]{2019Fake}
\begin{botherref}
\oauthor{\bsnm{Zhou}, \binits{X.}},
\oauthor{\bsnm{Jain}, \binits{A.}},
\oauthor{\bsnm{Phoha}, \binits{V.V.}},
\oauthor{\bsnm{Zafarani}, \binits{R.}}:
Fake news early detection: An interdisciplinary study.
arXiv e-prints
(2019)
\end{botherref}
\endbibitem

%%% 17
\bibitem[\protect\citeauthoryear{Chen and Guestrin}{2016}]{2016XGBoost}
\begin{bchapter}
\bauthor{\bsnm{Chen}, \binits{T.}},
\bauthor{\bsnm{Guestrin}, \binits{C.}}:
\bctitle{Xgboost: A scalable tree boosting system}.
In: \bbtitle{Knowledge Discovery and Data Mining}
(\byear{2016})
\end{bchapter}
\endbibitem

%%% 18
\bibitem[\protect\citeauthoryear{Kim}{2014}]{2014Convolutional}
\begin{botherref}
\oauthor{\bsnm{Kim}, \binits{Y.}}:
Convolutional neural networks for sentence classification.
Eprint Arxiv
(2014)
\end{botherref}
\endbibitem

%%% 19
\bibitem[\protect\citeauthoryear{Hochreiter and Schmidhuber}{1997}]{1997Long}
\begin{barticle}
\bauthor{\bsnm{Hochreiter}, \binits{S.}},
\bauthor{\bsnm{Schmidhuber}, \binits{J.}}:
\batitle{Long short-term memory}.
\bjtitle{Neural Computation}
\bvolume{9}(\bissue{8}),
\bfpage{1735}--\blpage{1780}
(\byear{1997})
\end{barticle}
\endbibitem

%%% 20
\bibitem[\protect\citeauthoryear{Cho et~al.}{2014}]{2014Learning}
\begin{botherref}
\oauthor{\bsnm{Cho}, \binits{K.}},
\oauthor{\bsnm{Merrienboer}, \binits{B.V.}},
\oauthor{\bsnm{Gulcehre}, \binits{C.}},
\oauthor{\bsnm{Bahdanau}, \binits{D.}},
\oauthor{\bsnm{Bougares}, \binits{F.}},
\oauthor{\bsnm{Schwenk}, \binits{H.}},
\oauthor{\bsnm{Bengio}, \binits{Y.}}:
Learning phrase representations using rnn encoder-decoder for statistical
  machine translation.
Association for Computational Linguistics
(2014)
\end{botherref}
\endbibitem

%%% 21
\bibitem[\protect\citeauthoryear{Devlin et~al.}{2018}]{2018BERT}
\begin{botherref}
\oauthor{\bsnm{Devlin}, \binits{J.}},
\oauthor{\bsnm{Chang}, \binits{M.W.}},
\oauthor{\bsnm{Lee}, \binits{K.}},
\oauthor{\bsnm{Toutanova}, \binits{K.}}:
Bert: Pre-training of deep bidirectional transformers for language
  understanding
(2018)
\end{botherref}
\endbibitem

%%% 22
\bibitem[\protect\citeauthoryear{Jain et~al.}{}]{0CanarDeep}
\begin{botherref}
\oauthor{\bsnm{Jain}, \binits{D.K.}},
\oauthor{\bsnm{Kumar}, \binits{A.}},
\oauthor{\bsnm{Shrivastava}, \binits{A.}}:
Canardeep: a hybrid deep neural model with mixed fusion for rumour detection in
  social data streams.
Neural Computing and Applications,
1--12
\end{botherref}
\endbibitem

%%% 23
\bibitem[\protect\citeauthoryear{Shi et~al.}{2023}]{2023PL}
\begin{botherref}
\oauthor{\bsnm{Shi}, \binits{Y.}},
\oauthor{\bsnm{Zhang}, \binits{X.}},
\oauthor{\bsnm{Yu}, \binits{N.}}:
Pl-transformer: a pos-aware and layer ensemble transformer for text
  classification.
Neural Computing and Applications,
1--12
(2023)
\end{botherref}
\endbibitem

%%% 24
\bibitem[\protect\citeauthoryear{Liu et~al.}{2019}]{liu2019roberta}
\begin{botherref}
\oauthor{\bsnm{Liu}, \binits{Y.}},
\oauthor{\bsnm{Ott}, \binits{M.}},
\oauthor{\bsnm{Goyal}, \binits{N.}},
\oauthor{\bsnm{Du}, \binits{J.}},
\oauthor{\bsnm{Joshi}, \binits{M.}},
\oauthor{\bsnm{Chen}, \binits{D.}},
\oauthor{\bsnm{Levy}, \binits{O.}},
\oauthor{\bsnm{Lewis}, \binits{M.}},
\oauthor{\bsnm{Zettlemoyer}, \binits{L.}},
\oauthor{\bsnm{Stoyanov}, \binits{V.}}:
Roberta: A robustly optimized bert pretraining approach.
arXiv preprint arXiv:1907.11692
(2019)
\end{botherref}
\endbibitem

%%% 25
\bibitem[\protect\citeauthoryear{Kula et~al.}{2021}]{kula2021implementation}
\begin{botherref}
\oauthor{\bsnm{Kula}, \binits{S.}},
\oauthor{\bsnm{Kozik}, \binits{R.}},
\oauthor{\bsnm{Chora{\'s}}, \binits{M.}}:
Implementation of the bert-derived architectures to tackle disinformation
  challenges.
Neural Computing and Applications,
1--13
(2021)
\end{botherref}
\endbibitem

%%% 26
\bibitem[\protect\citeauthoryear{Jain et~al.}{2022}]{jain2022aenet}
\begin{barticle}
\bauthor{\bsnm{Jain}, \binits{V.}},
\bauthor{\bsnm{Kaliyar}, \binits{R.K.}},
\bauthor{\bsnm{Goswami}, \binits{A.}},
\bauthor{\bsnm{Narang}, \binits{P.}},
\bauthor{\bsnm{Sharma}, \binits{Y.}}:
\batitle{Aenet: an attention-enabled neural architecture for fake news
  detection using contextual features}.
\bjtitle{Neural Computing and Applications}
\bvolume{34}(\bissue{1}),
\bfpage{771}--\blpage{782}
(\byear{2022})
\end{barticle}
\endbibitem

%%% 27
\bibitem[\protect\citeauthoryear{Liu et~al.}{2016}]{2016Recurrent}
\begin{botherref}
\oauthor{\bsnm{Liu}, \binits{P.}},
\oauthor{\bsnm{Qiu}, \binits{X.}},
\oauthor{\bsnm{Huang}, \binits{X.}}:
Recurrent Neural Network for Text Classification with Multi-Task Learning.
AAAI Press
(2016)
\end{botherref}
\endbibitem

%%% 28
\bibitem[\protect\citeauthoryear{Joulin et~al.}{2017}]{2017Bag}
\begin{botherref}
\oauthor{\bsnm{Joulin}, \binits{A.}},
\oauthor{\bsnm{Grave}, \binits{E.}},
\oauthor{\bsnm{Bojanowski}, \binits{P.}},
\oauthor{\bsnm{Mikolov}, \binits{T.}}:
Bag of tricks for efficient text classification
(2017)
\end{botherref}
\endbibitem

%%% 29
\bibitem[\protect\citeauthoryear{Lin}{2020}]{dxy-rumor}
\begin{botherref}
\oauthor{\bsnm{Lin}, \binits{I.}}:
(2020).
\url{http://hdl.handle.net/20.500.12675/sdp.164220}
\end{botherref}
\endbibitem

%%% 30
\bibitem[\protect\citeauthoryear{Pedregosa et~al.}{2011}]{pedregosa2011scikit}
\begin{barticle}
\bauthor{\bsnm{Pedregosa}, \binits{F.}},
\bauthor{\bsnm{Varoquaux}, \binits{G.}},
\bauthor{\bsnm{Gramfort}, \binits{A.}},
\bauthor{\bsnm{Michel}, \binits{V.}},
\bauthor{\bsnm{Thirion}, \binits{B.}},
\bauthor{\bsnm{Grisel}, \binits{O.}},
\bauthor{\bsnm{Blondel}, \binits{M.}},
\bauthor{\bsnm{Prettenhofer}, \binits{P.}},
\bauthor{\bsnm{Weiss}, \binits{R.}},
\bauthor{\bsnm{Dubourg}, \binits{V.}}, \betal:
\batitle{Scikit-learn: Machine learning in python}.
\bjtitle{the Journal of machine Learning research}
\bvolume{12},
\bfpage{2825}--\blpage{2830}
(\byear{2011})
\end{barticle}
\endbibitem

%%% 31
\bibitem[\protect\citeauthoryear{Wang et~al.}{2022}]{wang2022inductgcn}
\begin{botherref}
\oauthor{\bsnm{Wang}, \binits{K.}},
\oauthor{\bsnm{Han}, \binits{S.C.}},
\oauthor{\bsnm{Poon}, \binits{J.}}:
InducT-GCN: Inductive Graph Convolutional Networks for Text Classification
(2022)
\end{botherref}
\endbibitem

%%% 32
\bibitem[\protect\citeauthoryear{Lin et~al.}{2021}]{2021BertGCN}
\begin{botherref}
\oauthor{\bsnm{Lin}, \binits{Y.}},
\oauthor{\bsnm{Meng}, \binits{Y.}},
\oauthor{\bsnm{Sun}, \binits{X.}},
\oauthor{\bsnm{Han}, \binits{Q.}},
\oauthor{\bsnm{Wu}, \binits{F.}}:
Bertgcn: Transductive text classification by combining gnn and bert
(2021)
\end{botherref}
\endbibitem

%%% 33
\bibitem[\protect\citeauthoryear{He et~al.}{2021}]{he2021deberta}
\begin{botherref}
\oauthor{\bsnm{He}, \binits{P.}},
\oauthor{\bsnm{Liu}, \binits{X.}},
\oauthor{\bsnm{Gao}, \binits{J.}},
\oauthor{\bsnm{Chen}, \binits{W.}}:
DeBERTa: Decoding-enhanced BERT with Disentangled Attention
(2021)
\end{botherref}
\endbibitem

%%% 34
\bibitem[\protect\citeauthoryear{Johnson and Tong}{2017}]{2017Deep}
\begin{bchapter}
\bauthor{\bsnm{Johnson}, \binits{R.}},
\bauthor{\bsnm{Tong}, \binits{Z.}}:
\bctitle{Deep pyramid convolutional neural networks for text categorization}.
In: \bbtitle{Proceedings of the 55th Annual Meeting of the Association for
  Computational Linguistics (Volume 1: Long Papers)}
(\byear{2017})
\end{bchapter}
\endbibitem

%%% 35
\bibitem[\protect\citeauthoryear{Vaswani et~al.}{2017}]{2017attention}
\begin{botherref}
\oauthor{\bsnm{Vaswani}, \binits{A.}},
\oauthor{\bsnm{Shazeer}, \binits{N.}},
\oauthor{\bsnm{Parmar}, \binits{N.}},
\oauthor{\bsnm{Uszkoreit}, \binits{J.}},
\oauthor{\bsnm{Jones}, \binits{L.}},
\oauthor{\bsnm{Gomez}, \binits{A.N.}},
\oauthor{\bsnm{Kaiser}, \binits{L.}},
\oauthor{\bsnm{Polosukhin}, \binits{I.}}:
Attention is all you need.
arXiv
(2017)
\end{botherref}
\endbibitem

%%% 36
\bibitem[\protect\citeauthoryear{Peng et~al.}{2016}]{2016attention}
\begin{bchapter}
\bauthor{\bsnm{Peng}, \binits{Z.}},
\bauthor{\bsnm{Wei}, \binits{S.}},
\bauthor{\bsnm{Tian}, \binits{J.}},
\bauthor{\bsnm{Qi}, \binits{Z.}},
\bauthor{\bsnm{Bo}, \binits{X.}}:
\bctitle{Attention-based bidirectional long short-term memory networks for
  relation classification}.
In: \bbtitle{Proceedings of the 54th Annual Meeting of the Association for
  Computational Linguistics (Volume 2: Short Papers)}
(\byear{2016})
\end{bchapter}
\endbibitem

%%% 37
\bibitem[\protect\citeauthoryear{Lai et~al.}{2015}]{2015Recurrent}
\begin{bchapter}
\bauthor{\bsnm{Lai}, \binits{S.}},
\bauthor{\bsnm{Xu}, \binits{L.}},
\bauthor{\bsnm{Liu}, \binits{K.}},
\bauthor{\bsnm{Zhao}, \binits{J.}}:
\bctitle{Recurrent convolutional neural networks for text classification}.
In: \bbtitle{National Conference on Artificial Intelligence}
(\byear{2015})
\end{bchapter}
\endbibitem

%%% 38
\bibitem[\protect\citeauthoryear{Li et~al.}{2018}]{P18-2023}
\begin{bchapter}
\bauthor{\bsnm{Li}, \binits{S.}},
\bauthor{\bsnm{Zhao}, \binits{Z.}},
\bauthor{\bsnm{Hu}, \binits{R.}},
\bauthor{\bsnm{Li}, \binits{W.}},
\bauthor{\bsnm{Liu}, \binits{T.}},
\bauthor{\bsnm{Du}, \binits{X.}}:
\bctitle{Analogical reasoning on chinese morphological and semantic relations}.
In: \bbtitle{Proceedings of the 56th Annual Meeting of the Association for
  Computational Linguistics (Volume 2: Short Papers)},
pp. \bfpage{138}--\blpage{143}.
\bpublisher{Association for Computational Linguistics}, \blocation{???}
(\byear{2018}).
\burl{http://aclweb.org/anthology/P18-2023}
\end{bchapter}
\endbibitem

\end{thebibliography}
% \section*{Declarations}

% Some journals require declarations to be submitted in a standardised format. Please check the Instructions for Authors of the journal to which you are submitting to see if you need to complete this section. If yes, your manuscript must contain the following sections under the heading `Declarations':

% \begin{itemize}
% \item Conflict of interest/Competing interests (check journal-specific guidelines for which heading to use)
% \item Ethics approval 
% \item Consent to participate
% \item Consent for publication
% \item Availability of data and materials
% \item Authors' contributions
% \end{itemize}

% \noindent
% If any of the sections are not relevant to your manuscript, please include the heading and write `Not applicable' for that section. 

%%===================================================%%
%% For presentation purposes, we have included        %%

% Springer journals and proceedings: \url{https://www.springer.com/gp/editorial-policies}

% Nature Portfolio journals: \url{https://www.nature.com/nature-research/editorial-policies}

% \textit{Scientific Reports}: \url{https://www.nature.com/srep/journal-policies/editorial-policies}

% BMC journals: \url{https://www.biomedcentral.com/getpublished/editorial-policies}
% \end{flushleft}

\begin{appendices}

% \section{Section title of first appendix}\label{secA1}

% An appendix contains supplementary information that is not an essential part of the text itself but which may help provide a more comprehensive understanding of the research problem or it is information that is too cumbersome to be included in the body of the paper.




\end{appendices}


\end{document}